\documentclass{article} % For LaTeX2e
\usepackage[final]{colm2025_conference}

\usepackage{microtype}
\usepackage{hyperref}
\usepackage{url}
\usepackage{booktabs}

\usepackage{lineno}
\usepackage{amsmath}
\usepackage{graphicx}

\definecolor{darkblue}{rgb}{0, 0, 0.5}
\hypersetup{colorlinks=true, citecolor=darkblue, linkcolor=darkblue, urlcolor=darkblue}

\title{Generalized Graph Transformer Variational Autoencoder}

\author{Siddhant Karki \\
Department of Computer Science and Software Engineering \\
Miami University, Oxford, OH 45056, USA \\
\texttt{karkiss@miamioh.edu}
}

\begin{document}

\ifcolmsubmission
\linenumbers
\fi

\maketitle

\begin{abstract}
\end{abstract}
Graph link prediction has long been a central problem in graph representation learning in both network analysis and generative modeling. Recent progress in deep learning has introduced increasingly sophisticated architectures for capturing relational dependencies within graph-structured data. In this work, we propose the Generalized Graph Transformer Variational Autoencoder (GGT-VAE). Our model integrates Generalized Graph Transformer Architecture with Variational Autoencoder framework for link prediction. Unlike prior GraphVAE, GCN, or GNN approaches, GGT-VAE leverages transformer style global self-attention mechanism along with laplacian positional encoding to model structural patterns across nodes into a latent space without relying on message passing. Experimental results on several benchmark datasets demonstrate that GGT-VAE consistently achieves above-baseline performance in terms of ROC-AUC and Average Precision. To the best of our knowledge, this is among the first studies to explore graph structure generation using a generalized graph transformer backbone in a variational framework.
% \begin{abstract}
% Crowdsourced mapping platforms such as OpenStreetMap often contain incomplete or inconsistent information, including missing attributes and structural issues within polygonal map data. There has been limited research conducted on image-to-graph modeling to address this issue. In this work, we explore Graph Attention–based Variational Autoencoders (GT-VAE) designed for structural learning in spatial graphs. Our model demonstrates exceptional performance in link prediction tasks on road network graphs, achieving an average precision of 99\%, relying solely on the 2D positional features of nodes extracted from imagery. To the best of our knowledge, this is one of the first studies to explore graph structure prediction for real-world road networks, establishing strong baseline results for future research in this domain.
% \end{abstract}

\section{Introduction}
One of the most fundamental problems in graph machine learning is \textit{link prediction}, the task of inferring missing or potential edges between nodes. Link prediction enables applications such as recommending new friends in social media, predicting and modeling latent associations in molecular bonds in chemistry, and road networks in remote sensing.

Traditional graph learning methods relied on hand, engineered features, matrix factorization, or random walk–based embeddings such as DeepWalk and node2vec. While effective to some extent, these methods are limited in their ability to generalize structural information beyond observed connectivity patterns. The emergence of deep learning–based models for graphs has led to significant progress in modeling node and edge representations.

Early works like the Graph Convolutional Network (GCN) extended convolutional principles to graphs by aggregating neighborhood information. Building upon this, \textit{Variational Graph Autoencoders} (VGAEs)~\citep{kipf2016variationalgraphautoencoders} introduced a probabilistic approach to link prediction, combining graph convolutional encoders with variational inference to learn a latent distribution over graph structures. The objective can be formulated as maximizing the evidence lower bound (ELBO):

\[
\mathcal{L}_{VGAE} = \mathbb{E}_{q_\phi(\mathbf{Z}|\mathbf{X},\mathbf{A})} [\log p_\theta(\mathbf{A}|\mathbf{Z})] 
- D_{KL}\!\big(q_\phi(\mathbf{Z}|\mathbf{X},\mathbf{A}) \| p(\mathbf{Z})\big),
\]
where $\mathbf{Z}$ denotes the latent node embeddings parameterized by encoder parameters $\phi$.

Despite their success, GCN- and GAT-based approaches remain limited by their local message passing, which makes it difficult to capture long-range relationships in graphs. In contrast, \textit{transformer architectures}~\citep{vaswani2023attentionneed} introduce a global self-attention mechanism that directly models pairwise interactions between all elements, offering a natural way to learn global relationships. This idea led to graph-specific transformer models such as Graph-BERT~\citep{zhang2020graphbertattentionneededlearning}, Graph Transformer Networks (GTN)~\citep{yun2020graphtransformernetworks}, and LPFormer~\citep{Shomer_2024}, which apply attention to graph-structured data. While these models improve global context understanding, they generally work in a deterministic way and do not include a probabilistic latent space.

Recent work has started combining variational inference with transformers to enhance graph generation. For instance, GraphVAE~\citep{simonovsky2018graphvaegenerationsmallgraphs} and the Transformer Graph VAE for molecular generation~\citep{NGUYEN2025} show that combining these two frameworks can produce more expressive generative models. However, these methods mainly focus on small or domain-specific graphs and still depend on message-passing mechanisms from Graph Attention Networks.

Motivated by these developments, we explore whether a \textit{vanilla of graph transformer}~\citep{dwivedi2021generalizationtransformernetworksgraphs} , when used within a variational autoencoder, can serve as a strong generative model for graph structure prediction. Our model, the \textbf{Generalized Graph Transformer Variational Autoencoder (GGT-VAE)}, removes explicit message passing and instead uses Laplacian positional encodings with self-attentionto learn both local and global structure for link prediction. This work aims to bridge the gap between probabilistic graph modeling and transformer-inspired graph architectures.

\noindent \textbf{Our main contributions are as follows:}
\begin{itemize}
    \item We introduce a \textbf{Generalized Graph Transformer Variational Autoencoder (GGT-VAE)} that combines transformer-style self-attention with variational inference for link prediction.
    \item Our model captures both \textbf{local and global structure} without relying on message passing.
    \item Experimental results on benchmark datasets show that GGT-VAE achieves competitive or superior link prediction performance compared to message-passing VAEs.
\end{itemize}
\section{Related Work}

\subsection{Autoencoders and Variational Autoencoders}
Autoencoders (AEs) are fundamental generative models that learn to reconstruct inputs by mapping them into a lower-dimensional latent space. Formally, an encoder network $f_\phi(\mathbf{x})$ projects an input $\mathbf{x}$ to a latent representation $\mathbf{z}$, while a decoder $g_\theta(\mathbf{z})$ reconstructs the input as $\hat{\mathbf{x}} = g_\theta(f_\phi(\mathbf{x}))$. The objective minimizes a reconstruction loss:
\[
\mathcal{L}_{AE} = \|\mathbf{x} - \hat{\mathbf{x}}\|^2.
\]
Building upon this framework, \citet{kingma2022autoencodingvariationalbayes,Kingma_2019} introduced the \textit{Variational Autoencoder} (VAE), which imposes a probabilistic structure on the latent space by introducing an approximate posterior $q_\phi(\mathbf{z}|\mathbf{x})$ and prior $p(\mathbf{z})$. The model optimizes the evidence lower bound (ELBO):
\[
\mathcal{L}_{VAE} = \mathbb{E}_{q_\phi(\mathbf{z}|\mathbf{x})}[\log p_\theta(\mathbf{x}|\mathbf{z})] - D_{KL}\big(q_\phi(\mathbf{z}|\mathbf{x}) \| p(\mathbf{z})\big).
\]

\subsection{Graph Autoencoders and Graph Neural Networks}
\cite{kipf2016variationalgraphautoencoders} extended the VAE framework to graph-structured data with the \textit{Variational Graph Autoencoder} (VGAE). Here, node embeddings $\mathbf{Z}$ are learned via a Graph Convolutional Network (GCN) encoder, which aggregates neighborhood information as:
\[
\mathbf{Z} = \text{GCN}(\mathbf{X}, \mathbf{A}) = \sigma(\tilde{\mathbf{D}}^{-\frac{1}{2}}\tilde{\mathbf{A}}\tilde{\mathbf{D}}^{-\frac{1}{2}}\mathbf{X}\mathbf{W}),
\]
where $\mathbf{A}$ is the adjacency matrix and $\tilde{\mathbf{A}} = \mathbf{A} + \mathbf{I}$. The decoder reconstructs the graph via inner products, $\hat{\mathbf{A}} = \sigma(\mathbf{Z}\mathbf{Z}^\top)$, making VGAE a natural choice for link prediction.  

Subsequent work on \textit{Graph Attention Networks (GAT)}~\citep{veličković2018graphattentionnetworks} replaced fixed convolutional aggregation with learnable attention coefficients:
\[
\alpha_{ij} = \frac{\exp(\text{LeakyReLU}(\mathbf{a}^\top[\mathbf{W}\mathbf{h}_i \| \mathbf{W}\mathbf{h}_j]))}{\sum_{k \in \mathcal{N}(i)} \exp(\text{LeakyReLU}(\mathbf{a}^\top[\mathbf{W}\mathbf{h}_i \| \mathbf{W}\mathbf{h}_k]))},
\]
enabling more flexible, content-dependent message passing. Despite this, both GCN and GAT architectures are inherently local, relying on neighborhood-level propagation that limits their receptive field.

\subsection{Transformers for Graph Representation Learning}
Transformers~\citep{vaswani2023attentionneed} introduced the concept of global self-attention, where token interactions are modeled through the scaled dot-product:
\[
\text{Attention}(\mathbf{Q}, \mathbf{K}, \mathbf{V}) = \text{softmax}\!\left(\frac{\mathbf{Q}\mathbf{K}^\top}{\sqrt{d_k}}\right)\!\mathbf{V}.
\]
Originally proposed for language tasks, the mechanism generalizes naturally to graphs by treating nodes as tokens. Graph-BERT~\citep{zhang2020graphbertattentionneededlearning} applies this idea to graph data, replacing message passing with attention-based aggregation and positional encodings that capture global graph structure. Similarly, the Graph Transformer Network (GTN)~\citep{yun2020graphtransformernetworks} integrates attention over multiple edge types, while still retaining message-passing elements.

\subsection{Transformer Models for Link Prediction}
More recent work, such as LPFormer~\citep{Shomer_2024}, employs a pure transformer architecture for link prediction and relies on message passing neural network architecture (MPNNs). However, LPFormer remains a deterministic encoder–decoder that relies on MPNN for encoding. In addition, they lack the probabilistic latent structure of VAEs that supports graph generation.

\subsection{Graph Generation and Molecular Design}
Graph generation using VAE-based architectures has been explored by \citet{simonovsky2018graphvaegenerationsmallgraphs}, who introduced GraphVAE for generating small graphs. Later, \citet{NGUYEN2025} extended this idea by combining transformers with GAT-VAEs for molecular generation. Their model used cross-attention between latent embeddings and a SMILES text-transformer encoder to capture both local and long-range molecular dependencies. Together, these studies show that integrating transformer architectures with variational inference can effectively support graph-level generation tasks.

\vspace{0.5em}
These developments, from autoencoders to graph transformers, highlight a natural progression toward models that jointly capture global structural patterns and latent probabilistic representations, motivating the framework described in the next section.

In summary, autoencoders and their variational extensions provide the theoretical backbone for latent generative modeling. Transformers introduce relational reasoning through attention, while GCNs and GATs adapt this mechanism to graph-structured data. The Graph Variational Autoencoder unites these ideas, offering a framework for probabilistic graph generation. Our work builds upon this foundation by incorporating attention-based graph encoders for link prediction tasks.

% Testing Architecture diagram goes here

\section{Methods}
\begin{figure}[t]
  \centering
  \includegraphics[width=1\textwidth]{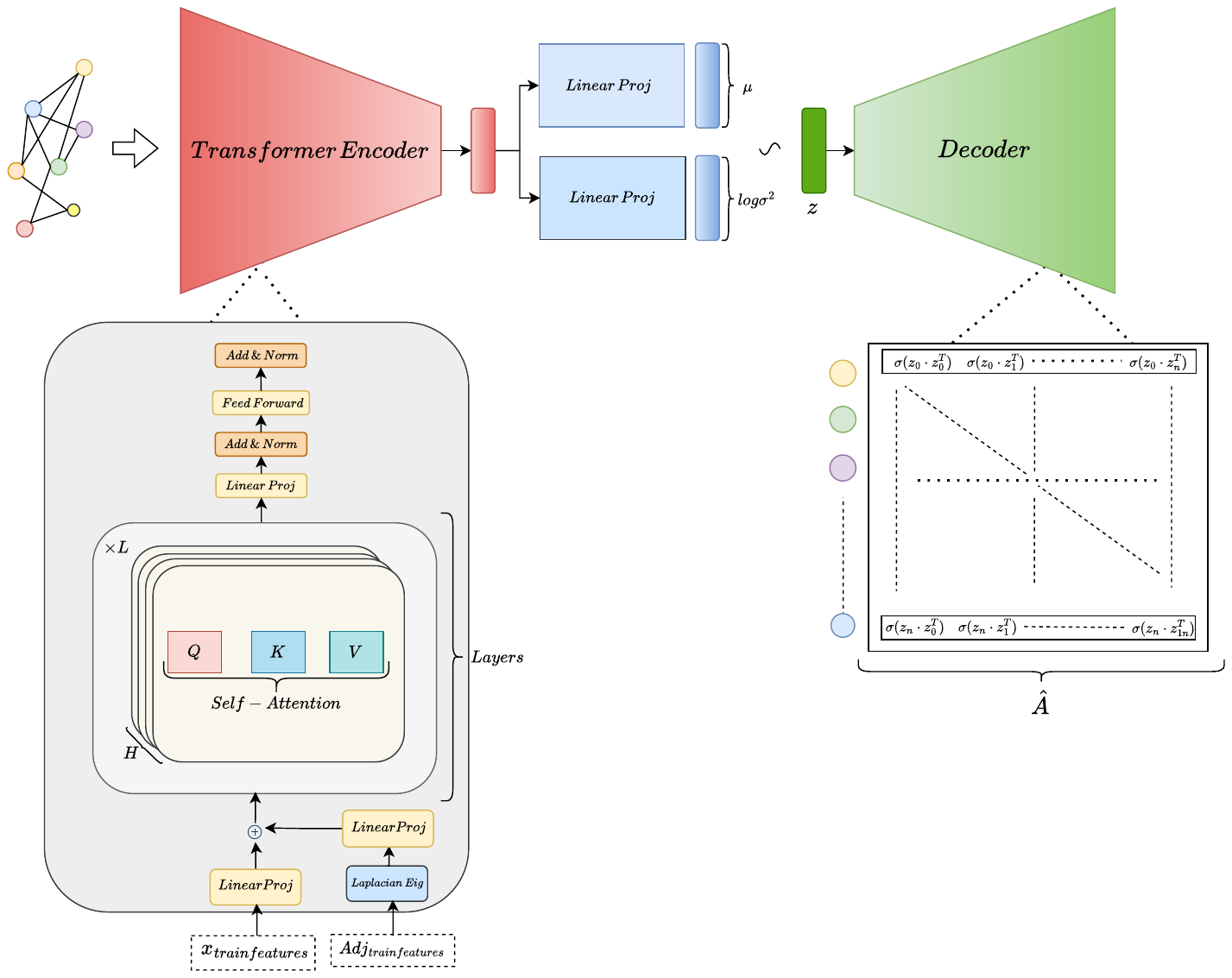}
  \caption{Architecture of the encoder–decoder framework used in our model. 
  The encoder maps node and positional embeddings into a latent space, 
  while the decoder reconstructs the adjacency matrix from the latent variables.}
  \label{fig:encoder_decoder}
\end{figure}
\label{Methods}
\vspace{0.5em}
Overall, the \textbf{GGT-VAE} follows a simple yet effective encoder–decoder structure. 
The encoder transforms node features and positional information into a compact latent space using self-attention, 
where each node’s representation depends on every other node in the graph. 
Unlike traditional GNNs, this removes the need for step-by-step neighborhood aggregation, 
allowing the model to learn global structure directly. 
The latent variables capture both local and global dependencies, while the decoder reconstructs the full graph by predicting the likelihood of edges between node pairs.

\subsection{Model Architecture}
We propose a \textbf{Generalized Graph Transformer Variational Autoencoder (GGT-VAE)} that models node dependencies without explicit message passing. 
Unlike GCN-based VAEs \citep{kipf2016variationalgraphautoencoders}, our encoder relies solely on transformer self-attention operating over laplacian position encoding produced from the training graph adjacency structure. 
This design allows the model to capture higher-order relationships while maintaining scalability and permutation invariance.

\paragraph{Input Embedding.}
Each node is represented by its raw feature vector $\mathbf{x}_i \in \mathbb{R}^{d_\text{node}}$ and positional encoding 
$\mathbf{p}_i \in \mathbb{R}^{d_\text{pos}}$, derived from the top-$k$ Laplacian eigenvectors.
The embedding module projects both components into a common latent space:
\begin{equation}
\mathbf{h}_i^{(0)} = \mathbf{W}_x \mathbf{x}_i + \mathbf{W}_p \mathbf{p}_i,
\end{equation}
where $\mathbf{W}_x, \mathbf{W}_p \in \mathbb{R}^{d_\text{hid} \times d_\text{in}}$ are learnable matrices. 
This yields an initial node representation $\mathbf{H}^{(0)} \in \mathbb{R}^{N \times d_\text{hid}}$.

\paragraph{Graph Transformer Encoder :} The encoder comprises $L$ stacked \textbf{Graph Transformer layers}, each containing $H$ self-attention heads. 
Given node representation at layer \(l\), $\mathbf{N}^{(l-1)} \in \mathbb{R}^{N \times d_\text{hid}}$ 
each head computes query, key, and value projections:
\begin{equation}
\mathbf{Q}_i = \mathbf{N}_i^{l-1}\mathbf{W}_Q, \quad
\mathbf{K}_i = \mathbf{N}_i^{l-1}\mathbf{W}_K, \quad
\mathbf{V}_i = \mathbf{N}_i^{l-1}\mathbf{W}_V.
\end{equation}
Attention weights are computed as scaled dot-products:
\begin{equation}
\mathbf{A}_i^\prime = \mathrm{softmax}\!\left( \frac{\mathbf{Q}\mathbf{K}^\top}{\sqrt{d_k}} \right),
\end{equation}
Outputs from all heads are concatenated and linearly projected:
\begin{equation}
\mathbf{N}^{{l-1}^{\prime}} = \text{Concat}(\mathbf{N}_1, \dots, \mathbf{N}_H)\mathbf{W}_O,
\end{equation}
followed by a residual connection, layer normalization, and a position-wise feedforward network:
\[
\mathbf{N}^{(L)} = \text{LayerNorm}\big(\mathbf{N}^{(l-1)} + \text{FFN}(\text{LayerNorm}(\mathbf{N}^{(l-1)} + \mathbf{N}^{{l-1}^{\prime}}))\big).
\]

\paragraph{Variational Encoding :} The encoder’s final output $\mathbf{N}^{(L)}$ is mapped to the parameters of a Gaussian latent distribution:
\[
\boldsymbol{\mu} = \mathbf{N}^{(L)} \mathbf{W}_\mu, \qquad
\log \boldsymbol{\sigma}^2 = \mathbf{N}^{(L)} \mathbf{W}_{\log{\sigma}^2}.
\]
A latent sample $\mathbf{Z}$ is obtained via the reparameterization trick:
\[
\mathbf{Z} = \boldsymbol{\mu} + \boldsymbol{\epsilon} \odot \exp(0.5\,\log \boldsymbol{\sigma}^2),
\quad \boldsymbol{\epsilon} \sim \mathcal{N}(0, I).
\]

\paragraph{Decoder :} Link probabilities are reconstructed through a simple inner-product decoder:
\[
\hat{\mathbf{A}} = \sigma(\mathbf{Z}\mathbf{Z}^\top),
\]
where $\sigma(\cdot)$ is the elementwise sigmoid function. 
Self-loops are masked to avoid trivial identity edges.

\paragraph{Training Objective :} The model is trained using the standard VAE loss:
\[
\mathcal{L} = \mathcal{L}_\text{recon} + \beta \, \mathcal{L}_\text{KL},
\]
where $\mathcal{L}_\text{recon}$ is the binary cross-entropy between $\hat{\mathbf{A}}$ and $\mathbf{A}$, 
and $\mathcal{L}_\text{KL}$ is the Kullback–Leibler divergence between the approximate posterior $q(\mathbf{Z}|\mathbf{X})$ 
and the unit Gaussian prior $p(\mathbf{Z})$.

GGT-VAE achieves comparable link prediction accuracy to message-passing VGAEs, 
while eliminating neighborhood aggregation. The attention mechanism allows flexible context modeling, 
and Laplacian-based positional encodings inject graph topology directly into the latent space.

\subsection{Data Collection \& Training Setup}

We use two standard citation network datasets: \textbf{Cora} and \textbf{Citeseer}.
They are downloaded from the Planetoid repository using \texttt{torch\_geometric.datasets.Planetoid}. 
After loading, each dataset is converted into a dense tensor format to make the model easier to run in PyTorch. 
This approach removes the need for sparse operations and makes the training setup simple and easy to reproduce.

For each dataset, we randomly split the edges into training, validation, and test sets. 
Because these graphs are very sparse, we balance the number of positive (real) and negative (false) edges during both training and evaluation. 
This helps the model learn fairly and prevents bias toward missing edges. 
During testing, we sample a smaller number of true and false edges than the maximum possible. 
This setup helps test the model’s ability to generalize beyond the exact edge patterns seen during training. 
The reconstruction loss is calculated using these balanced edge samples.

We train all models with the \textbf{AdamW optimizer}, a learning rate of $1\times10^{-3}$, and decay of $5\times10^{-4}$ and full-batch updates (the entire graph per step). 
All experiments were run on a single GPU, and random seeds were fixed for reproducibility.

\subsection{Training Configuration}

We trained multiple configurations of the model to study how different parameters affect performance. 
The number of transformer layers, attention heads, and hidden dimensions were varied to find a balance between accuracy and model complexity. 
We also tested different values of the $\beta$ term in the KL divergence to control the strength of regularization in the variational loss. All models were trained using the same setup described in Section~\ref{Methods}. 
The best configuration for each dataset was selected based on validation performance.
Results for the parameter variations are summarized in the next sections (see Table ~\ref{tab:abelations}) for details.

\section{Experimental Evaluation and Result Analysis}
\begin{figure*}[t]
  \centering
  % --- Attention Head Visualization ---
  \includegraphics[width=0.85\textwidth]{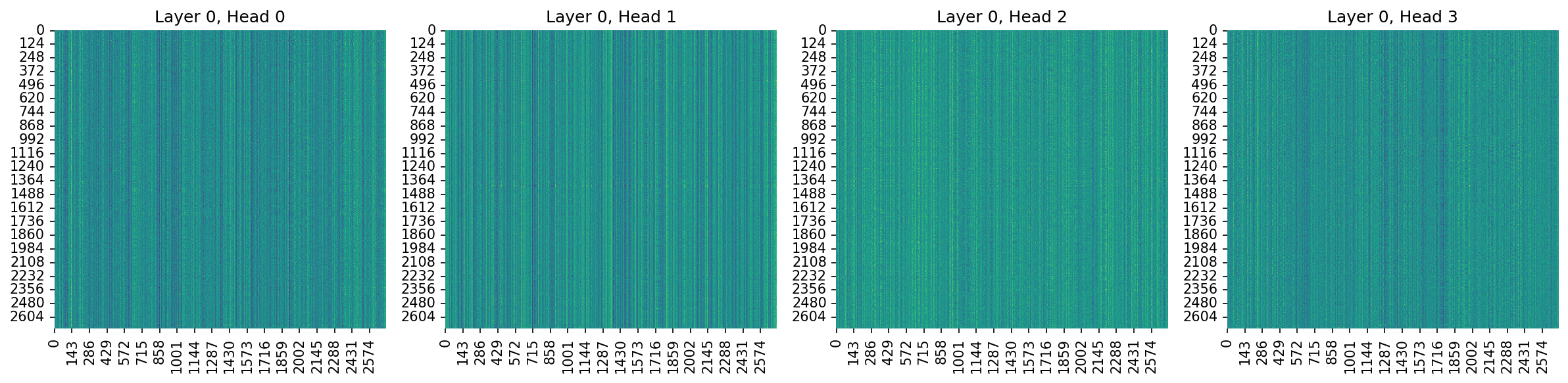}
  \hfill
  \includegraphics[width=0.85\textwidth]{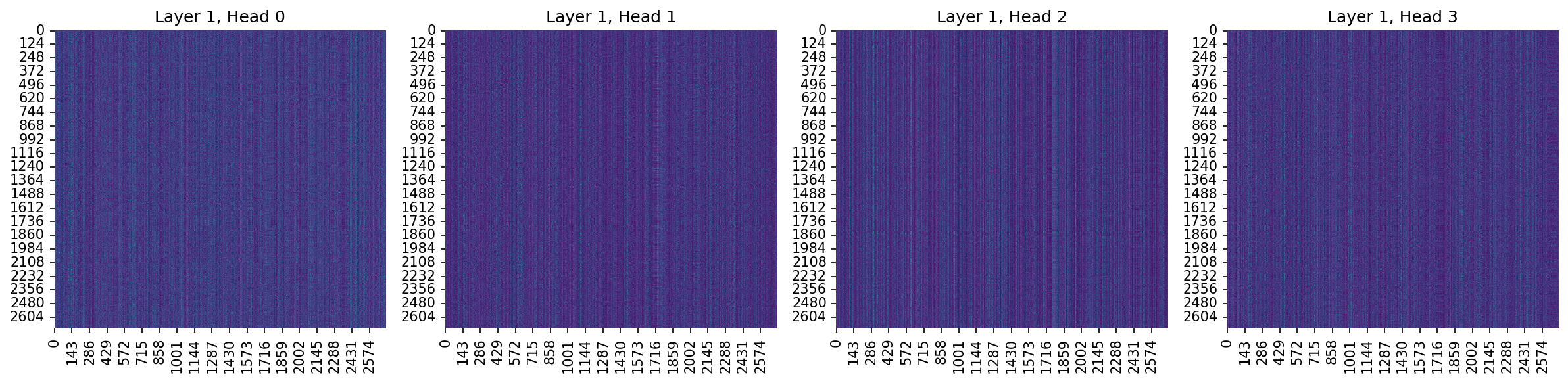}
  \hfill
  \includegraphics[width=0.85\textwidth]{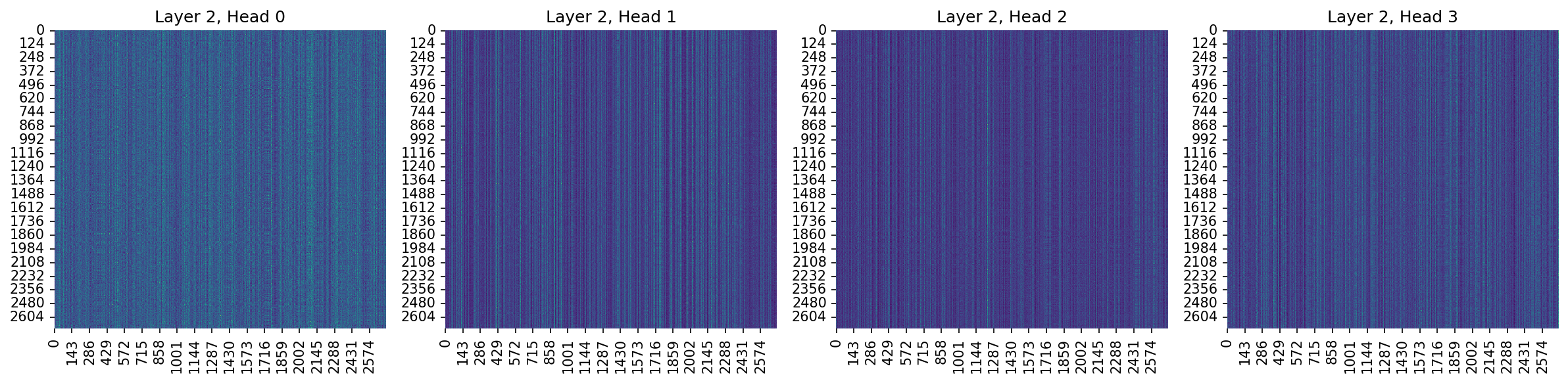}
  \includegraphics[width=0.85\textwidth]{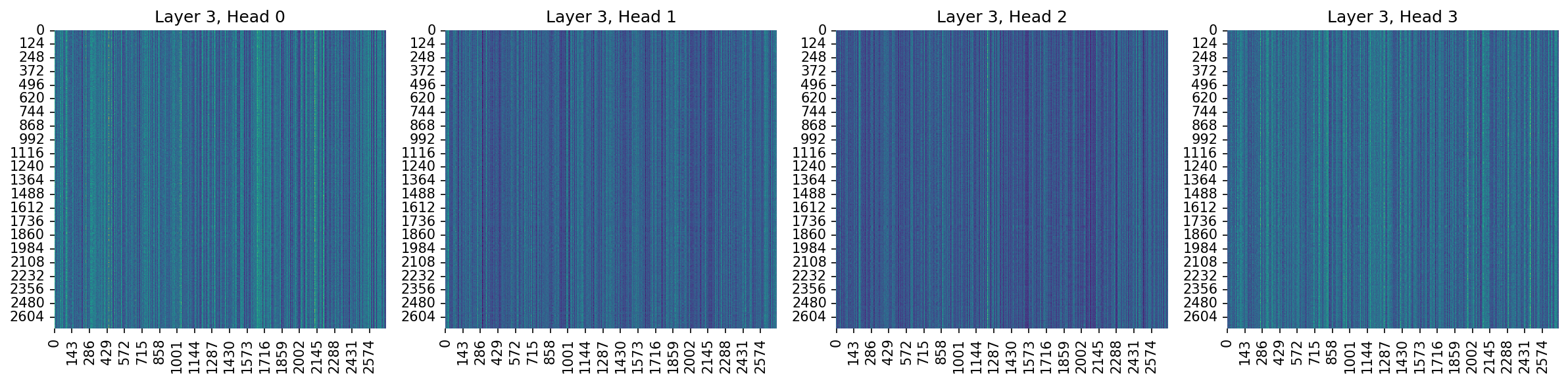}
  \caption{Attention maps for selected heads across Transformer layers. 
  Lighter regions correspond to stronger attention weights between node pairs. 
  Early layers focus on local neighborhoods, while deeper layers capture more global structural dependencies.}
  \label{fig:attention_maps}
\end{figure*}

\begin{figure}[t]
  \centering
  % --- t-SNE Visualization ---
  \includegraphics[width=0.5\textwidth]{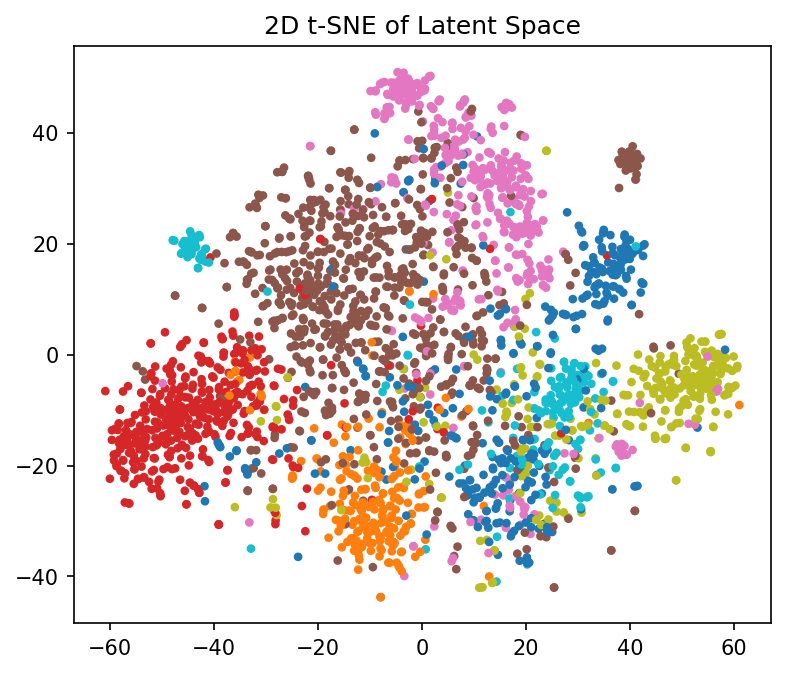}
  \caption{Cora dataset 2D t-SNE projection of latent embeddings from the encoder. 
  Nodes belonging to similar communities are clustered together, indicating the model’s ability to capture meaningful relational structure in the latent space.}
  \label{fig:tsne_latent}
\end{figure}

\subsection{Qualitative Analysis}
Figure~\ref{fig:attention_maps} visualizes the attention weights across transformer layers and heads. 
Each map shows how much one node attends to another, with lighter areas representing stronger attention values. 
A clear layer-by-layer progression can be observed. In \textbf{Layer 0}, the attention maps appear mostly uniform with low contrast, meaning that the model distributes attention evenly across nodes. 
This suggests that the first layer focuses on capturing graph-level information from laplacian positional encodings rather than detailed node level information. In \textbf{Layer 1}, faint vertical streaks begin to appear. 
These patterns indicate that certain nodes start receiving more global attention, often corresponding to central or highly connected nodes. 
The model begins recognizing the relative importance of different nodes in the graph. By \textbf{Layer 2}, the attention becomes more distinct, with some heads showing strong localized focus.
This shows that different heads specialize: some capture fine-grained, local structures, while others maintain the global awareness. 
Such diversity reflects the model’s ability to learn both local and global patterns simultaneously. In \textbf{Layer 3}, the maps display a mix of concentrated and uniform regions. 
Some heads show sharp vertical bands, focusing on specific node clusters, while others remain evenly distributed. 
This final layer balances global aggregation and local refinement, integrating information from previous layers into cohesive structural representations.

Across all layers, the variation among attention heads shows that the model naturally learns to combine local and global features.
This enables our model to capture structural information without relying on explicit message passing.
Figure~\ref{fig:tsne_latent} presents a 2D t-SNE visualization of the latent embeddings for \textbf{Cora} dataset. 
Nodes (ML papers) belonging to similar classes (types) cluster together, showing that the latent space preserves meaningful relationships between nodes. 
This clustering indicates that the encoder successfully learns a representation where the latent space reflects structural similarity in the graph.
\subsection{Quantitative Analysis}
\label{sec:analysis-nmp}
\begin{figure}[t]
  \centering
  \includegraphics[width=\linewidth]{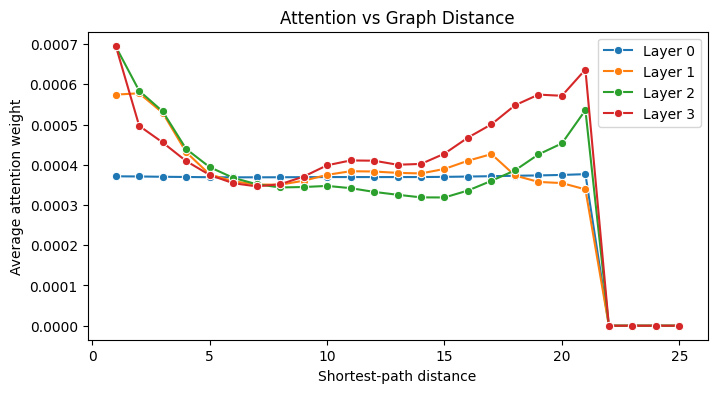}
  \vspace{-0.5em}
  \caption{\textbf{Attention vs.\ Graph Distance.}
  Average attention weight for different shortest-path distances (SPD) on Cora.
  Unlike message-passing models that only attend to direct neighbors ($\mathrm{SPD}=1$),
  the Transformer assigns nonzero weight even to distant nodes ($\mathrm{SPD}>10$),
  showing that it captures both local and global structure without relying on adjacency-based message passing.}
  \label{fig:attn-spd}
\end{figure}

\begin{figure}[t]
  \centering
  \includegraphics[width=0.8\linewidth]{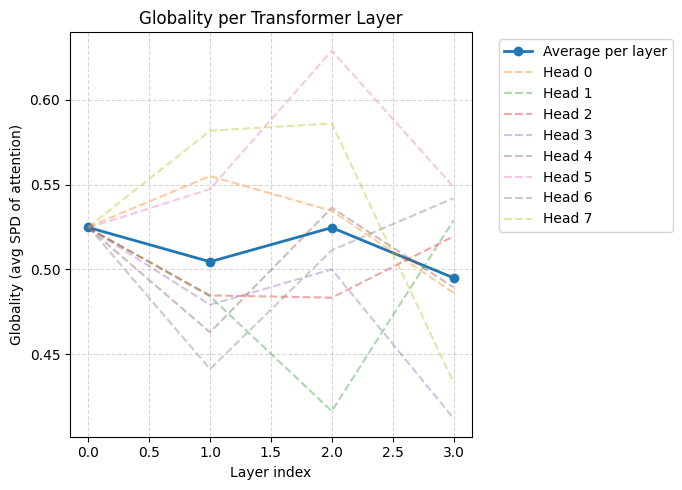}
  \vspace{-0.5em}
  \caption{\textbf{Normalized globality across layers and heads.}
  The solid line shows the layer average; dashed lines show individual heads.
  Globality rises in early layers (broader attention) and decreases in the final layer (local refinement).
  Diverging head trends show that some heads specialize in global structure while others focus on local details.}
  \label{fig:globality}
\end{figure}

In this section, we study how  learns graph structure without relying on message passing.  
To do this, we look at how attention spreads across different distances in the graph and how each layer’s focus changes as the model gets deeper.

\paragraph{Globality Metric.}
To understand how far each layer looks across the graph, we measure a value we call \emph{globality}.
It tells us the average distance (in graph hops) between nodes that attend to each other.  
For every layer $\ell$ and attention head $h$, we compute:
\begin{equation}
\text{Globality}_{\ell,h}
=
\frac{\sum_{d} d \cdot \bar{\alpha}^{(\ell,h)}(d)}
     {\sum_{d} \bar{\alpha}^{(\ell,h)}(d)},
\end{equation}
where $d$ is the shortest-path distance (SPD) between two nodes and $\bar{\alpha}^{(\ell,h)}(d)$ is the average attention weight between nodes at that distance.  
A smaller value means the model mostly attends to close neighbors (local attention), while a larger value means it also considers far-away nodes (global attention).

We also divide this score by the graph’s diameter to keep the values consistent between 0 and 1:
\begin{equation}
\text{NormalizedGlobality}_{\ell,h}
=
\frac{\text{Globality}_{\ell,h}}{D_{\text{graph}}}.
\end{equation}

\paragraph{Attention–Distance Analysis.}
Figure~\ref{fig:attn-spd} shows the average attention weight for different graph distances on the Cora dataset.  
In a message-passing network, information only moves between direct neighbors (\( \text{SPD}=1 \)).  
In contrast, GGT-VAE gives meaningful attention to nodes that are many hops apart (\( \text{SPD}>10 \)).  
The first layer shows almost uniform attention across all distances, meaning it starts with a global view.  
Middle layers balance both local (1–2 hop) and long-range connections, while the deepest layer again highlights both very near and very distant nodes.  
This pattern shows that the model learns to mix local and global information on its own, without needing any explicit neighborhood-based message passing.

\paragraph{Globality Across Layers and Heads.}
To understand how each layer and head contribute to the model’s receptive field, 
we measure the \emph{globality} of attention, the average distance between nodes that attend to each other, normalized by the graph’s diameter.
Figure~\ref{fig:globality} shows these values for all eight attention heads and the layer-wise average.

On average, globality increases slightly from Layer~0 to Layer~2, meaning the model initially expands its view to include more distant nodes.
In the final layer, the average drops again, showing that attention becomes somewhat more local once global information has been collected.
This pattern suggests a natural progression: early layers mix information broadly, while later layers refine relationships within smaller neighborhoods.

Across individual heads, we observe diverging trends.
Some heads consistently show higher globality, focusing on long-range structure and global context.
Others decrease sharply, concentrating on nearby nodes and preserving local detail.
This diversity across heads indicates that different parts of the Transformer specialize in complementary roles—
some capturing the big picture of the graph, others refining fine-grained connections.
Together, they enable the model to balance both local and global understanding without relying on message passing.

This confirms that the GGT-VAE learns to represent both local and global graph patterns without \emph{explicit message passing}.  
The model builds a latent space that reflects graph connectivity directly through attention, rather than by repeatedly aggregating features over edges. This explains why it performs well on link prediction: it can connect distant but structurally related nodes while keeping local information intact.

\subsection{Experimental Setup}

\begin{table}[t]
\centering
\small
\setlength{\tabcolsep}{5pt}
\caption{Link prediction on Planetoid datasets. Results are ROC-AUC / AP (\%). Higher is better.}
\label{tab:benchmark-results}
\begin{tabular}{lcc}
\toprule
\textbf{Method} & \textbf{Cora} & \textbf{Citeseer} \\
\midrule
DeepWalk \cite{stabilizing2022}              & 76.41 / 81.78 & 64.15 / 74.60 \\
Spectral Clustering \cite{stabilizing2022}   & 85.36 / 88.41 & 77.26 / 80.24 \\
Graphite-AE \cite{stabilizing2022}          & 91.29 / 92.63 & 88.84 / 89.64 \\
Linear-GAE \cite{stabilizing2022}           & 91.86 / 93.20 & 90.89 / 92.24 \\
2-GAE \cite{stabilizing2022}                & 91.61 / 92.81 & 89.04 / 89.26 \\
6-GAE \cite{stabilizing2022}                & 83.58 / 85.32 & 79.95 / 83.63 \\
6-DGAE$_{\text{a}}$ \cite{stabilizing2022}  & \underline{93.55} / \underline{94.46} & \underline{94.16} / \underline{94.86} \\
ARGA \cite{stabilizing2022}                 & 92.10 / 93.25 & 90.43 / 92.04 \\
\midrule
\textbf{GGT-VAE (ours)}                     & \textbf{92.04 $\pm$ 0.60} / \textbf{92.66 $\pm$ 0.80} & \textbf{92.00 $\pm$ 0.20} / \textbf{93.74 $\pm$ 0.16} \\
\bottomrule
\end{tabular}
\vspace{0.5mm}
\footnotesize{Results reproduced from \cite{stabilizing2022}. Our transformer-based VAE matches early VGAE-style baselines without message passing.}
\end{table}

We evaluate the proposed \textbf{GGT-VAE} on the standard Planetoid datasets \textbf{Cora} and \textbf{Citeseer} for the \textbf{link prediction} task. We follow the data splits and protocol of \cite{zhu2022graphvae} for comparability. Node features are used as-is, and Laplacian positional encodings supply structural information. The adjacency matrix for training is constructed only from training edges to prevent leakage. We train for 500 epochs with Adam (lr $=1\times 10^{-3}$), and apply early stopping with patience 50 on validation ROC-AUC. Unless noted, we use 4 transformer layers, 4 heads, hidden size 128, and $\beta = 0.5\times 10^{-3}$. All numbers are averaged over 10 random seeds and reported as mean $\pm$ std.

\subsection{Model Performance}

\begin{table}[t]
\centering
\small
\setlength{\tabcolsep}{3pt}
\caption{Ablations on \textbf{Cora}: effect of heads, layers, and hidden size. $\beta$ and LR are shown as $\beta$/LR. Results are ROC-AUC / AP (\%).}
\label{tab:ablations}
\begin{tabular}{lccccc}
\toprule
\textbf{Config} & \textbf{Heads} & \textbf{Layers} & \textbf{Hidden} & $\boldsymbol{\beta}$/LR & \textbf{Cora} \\
\midrule
Base  & 4 & 4 & 128 & $0.5\!\times\!10^{-3}$ / $1\!\times\!10^{-3}$ & 92.04 / 92.66 \\
\midrule
\multicolumn{6}{l}{\textit{Vary Heads (Layers=4, Dim=128)}} \\
H-1   & 1 & 4 & 128 & $0.5\!\times\!10^{-3}$ / $1\!\times\!10^{-3}$ & 90.89 / 92.44 \\
H-2   & 2 & 4 & 128 & $0.5\!\times\!10^{-3}$ / $1\!\times\!10^{-3}$ & 90.66 / 92.12 \\
H-8   & 8 & 4 & 128 & $0.5\!\times\!10^{-3}$ / $1\!\times\!10^{-3}$ & 91.03 / 92.49 \\
\midrule
\multicolumn{6}{l}{\textit{Vary Layers (Heads=4, Dim=128)}} \\
L-2   & 4 & 2 & 128 & $0.5\!\times\!10^{-3}$ / $1\!\times\!10^{-3}$ & 90.06 / 92.03 \\
L-8   & 4 & 8 & 128 & $0.5\!\times\!10^{-3}$ / $1\!\times\!10^{-5}$ & 48.88 / 49.94 \\
\midrule
\multicolumn{6}{l}{\textit{Vary Hidden Dim (Heads=4, Layers=4)}} \\
D-064 & 4 & 4 &  64 & $0.5\!\times\!10^{-3}$ / $1\!\times\!10^{-3}$ & 89.73 / 91.49 \\
D-256 & 4 & 4 & 256 & $0.5\!\times\!10^{-3}$ / $1\!\times\!10^{-5}$ & \textbf{92.23 / 93.81} \\
D-512 & 4 & 4 & 512 & $0.5\!\times\!10^{-3}$ / $1\!\times\!10^{-6}$ & 77.32 / 70.81 \\
\bottomrule
\end{tabular}
\end{table}

Table~\ref{tab:benchmark-results} shows that classic embedding methods (DeepWalk, Spectral Clustering) underperform because they lack a generative latent space. Message-passing autoencoders (Graphite-AE, Linear-GAE, ARGA) perform strongly by aggregating neighborhood information. Our \textbf{GGT-VAE} reaches \(92.04\%\) ROC-AUC / \(92.66\%\) AP on Cora and \(92.00\%\) ROC-AUC / \(93.74\%\) AP on Citeseer, demonstrating that transformer self-attention with structural encodings matches message-passing VAEs even without adjacency-based propagation.

\paragraph{Ablation studies.}
Table~\ref{tab:ablations} studies depth, heads, hidden size, and $\beta$. Four layers and four heads give the best stability. Fewer heads slightly hurt performance; more (8) does not yield further gains. Increasing hidden size to 256 helps, but very large hidden sizes (512) destabilize training. A moderate KL weight ($0.5\times 10^{-3}$) balances reconstruction and regularization. Overall, GGT-VAE is robust to moderate hyperparameter changes.

\paragraph{Summary.}
GGT-VAE attains high link-prediction accuracy on Cora and Citeseer using only self-attention and positional encodings. During testing we evaluate on a subsampled set of positive and negative edges to check generalization; the model maintains strong performance, indicating it does not overfit the training graph. This supports our claim that attention + variational inference is sufficient for graph representation learning without explicit message passing.

\section{Conclusion \& Future Work}

This paper introduced the \textbf{Generalized Graph Transformer Variational Autoencoder (GGT-VAE)} for link prediction. 
The model replaces traditional message passing with transformer self-attention and Laplacian positional encodings, allowing it to learn both local and global structure directly from the graph. 
Experiments on benchmark datasets show that GGT-VAE performs competitively with message-passing approaches while maintaining a simpler and more flexible architecture. 
Our analysis further shows that the model learns structural information naturally across layers, confirming that self-attention alone can effectively capture graph relationships.

Currently, the model focuses on edge generation from latent representations. 
Future work will extend this approach to full graph generation for a fixed number of nodes, incorporating node features as inputs. 
We plan to explore applications in molecular generation, where the model could learn complex chemical bonds and topologies, and in geospatial domains, such as generating road networks from satellite imagery. 
Another promising direction is conditioning graph generation on other modalities, such as images through cross-attention with vision transformers, or molecular SMILES text for chemistry tasks. 
These directions highlight the potential of combining graph transformers and variational autoencoders for broader multimodal and scientific applications.

\paragraph{Code Availability.}
To support transparency and reproducibility, we will release the full implementation, training scripts, and pre-trained model weights of GGT-VAE upon publication of the final version of this paper.
The codebase will include utilities for dataset preparation, model configuration, and visualization of attention and globality metrics to facilitate future research on transformer-based graph generative models.

\bibliography{colm2025_conference}
\bibliographystyle{colm2025_conference}
%\appendix
% \section{Appendix}
% You may include other additional sections here.
\end{document}